\documentclass[letterpaper]{article}
\usepackage{graphicx}
\usepackage{amssymb}
\usepackage{amsmath}
\usepackage{times}
\usepackage{hyperref}
\author{Christopher O. Ward\\Department of Mathematics and Computer Science\\University of West Indies (St. Augustine),\\
	Trinidad \& Tobago\\West Indies\\ \emph{christopher.ward@sta.uwi.edu}}
\title{BETTER GLOBAL POLYNOMIAL APPROXIMATION FOR IMAGE RECTIFICATION\footnote{The original Paper entitled "Better Global Polynomial Approximation for Image Rectification", was published in the International Journal of Modelling and Simulation, Vol. 28, No. 3, 2008, pp 299-308.}}

\newtheorem{theorem}{Theorem}
\newtheorem{definition}{Definition}

\begin{document}

\maketitle

\begin{abstract}\noindent
When using images to locate objects, there is the problem of correcting for distortion and misalignment
in the images. An elegant way of solving this problem is to generate an error correcting function that
maps points in an image to their corrected locations. We generate such a function by fitting a polynomial
to a set of sample points. The objective is to identify a polynomial that passes ``sufficiently close'' to
these points with ``good'' approximation of intermediate points. In the past, it has been difficult to
achieve good global polynomial approximation using only sample points. We report on the development
of a global polynomial approximation algorithm for solving this problem.
\end{abstract}

\section{Introduction}

\begin{figure}[!htb]
 	\begin{tabular}{cc}
 	\includegraphics[width = 0.5\textwidth]{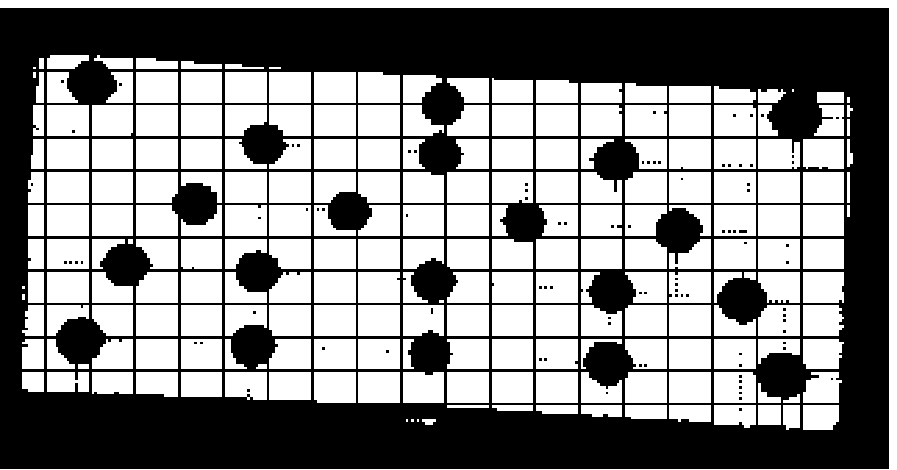} & \includegraphics[width = 0.5\textwidth]{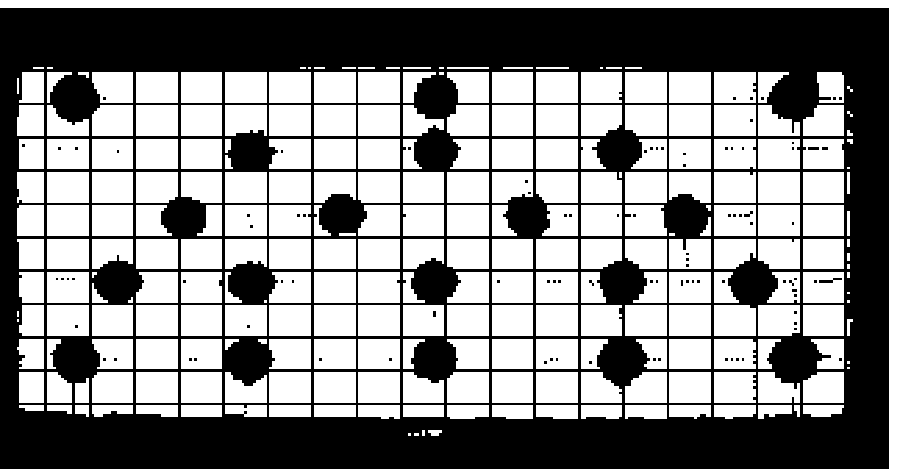}\\
 	{\small (a) unrectified image} & {\small (b) rectified image}
  	\end{tabular}
\caption{\small Rectification of an image of a test pattern placed over the robot's workspace taken from a severely misaligned camera. A grid has been superimposed on the images to aid in comparing the horizontal and vertical alignment of image features. Although the original images are colour coded, the images shown are monochrome. If the upper left dot and the upper right dot are ignored, the rest of the dots in (b) sample the area of the workspace that is within reach of the robot.}\label{fig:misaligned}
 \end{figure}

\noindent The problem that is addressed here occurred in the context of the development of a simple, low-cost
robotic exhibit for demonstrating the concept of intelligent robotics to the general public. Intelligent
robotics deals with the use of sensors to enhance a robot's performance in an uncertain environment. For the exhibit, we implemented a visually-guided pick-and-place robot. The robot uses an image to determine the location of objects placed arbitrarily on a flat surface and demonstrates success in locating the objects by manipulating them. The exhibit consists of a robotic arm that is fixed in front of a flat work surface. Two pedestals are placed anywhere within reach of the arm. One pedestal is blue and the other is green. A blue ball is placed on the blue pedestal and the robot must pick up the ball and place it on the green pedestal. An ordinary webcam is placed in a frame above the work surface and is used to determine the location of the pedestals so that the arm can be guided accordingly. Fig. \ref{fig:misaligned}(a) shows the camera's view of the workspace. Once the exhibit has been set up, camera, robot and workspace are all fixed relative to one another.

There is therefore the problem of determining the location of the colour-coded pedestals based on their image. This problem is compounded by the fact that the {webcam} produces significant image distortion, and by the fact that the position and orientation of the camera relative to the work surface may not be exactly the same every time the exhibit is set up. A similar problem occurs when any automated mechanism is taken apart for maintenance. The mechanism usually must be recalibrated when it is reassembled. In this sense, we are addressing the problem of easy recalibration of the vision component of our robotic exhibit.

In processing the image, there is a need to rectify significant image distortion caused by the optical properties of the camera and by errors in positioning the camera during set up of the exhibit. We solve this problem by determining a mapping from the pixel locations of image points to the physical locations of corresponding source points. The mapping has to be determined empirically in order to recalibrate the vision system each time the apparatus is set up. The pixel positions of the images of key points in a test
pattern are mapped to the known locations of these key points. This partial mapping is then used to approximate a mapping for the entire image. 

In  \cite{brown92survey}, Brown surveyed and classified several established methods for determining the mapping from pixel position to
source location for a camera. Using Brown's classification, the errors caused by distortion and
misalignment are static in the sense that they do not change from image to image taken with the same
camera in the same position. Static distortions can be rectified in a one-time-only setup process via calibration
techniques. Our method may be regarded as a calibration technique.

Brown classifies the distortion due to camera properties as internal and the misalignment as
external. Our approach does not require the use of any sort of model of the characteristics of the camera
or the geometry of how an image is captured. We treat the mechanism by which source points are captured as image points as a ``black box'', the inner workings of which is not modelled. We therefore handle these internal and external distortions
without having to distinguish between them.

Furthermore, we are concerned with geometric distortions as distinct from photometric
distortions. The effect of photometric distortions are minimised by using primary colours (red, green, blue) for the key features in the scene and either black or white for the other features. Features can
therefore be identified by colour without regard for intensity values. Colour distortions are handled by
filtering out colours below a fixed saturation value using a median filter to remove speckles and
discretising the remaining colours to fully saturated red, green or blue \cite{whelan2001}. Visual artifacts are then
located by looking for regions within the image with the appropriate colour.

The problem is thus reduced to one of finding an adequate approximation of a total mapping
from pixel positions to object locations using a set of sample points. Three popular approaches to
solving this problem are: the use of an artificial neural network; the use of a piecewise interpolation
technique to fit curves of a chosen form in between the sample points; the use of polynomial
interpolation to fit a polynomial to the sample points.

Artificial neural networks are popular among engineers for their ability to generalise from
sample data. We find it difficult to argue on practical grounds against this approach. However, for us,
the main difficulty with this method is its inability to yield a representation of the approximating
function that is independent from that of the neural network itself. However, recent advances in the use
of algebraic training of neural networks to produce closed form analytic solutions are addressing this
problem \cite{lagaris1997,ferrari2005}.

Piecewise interpolation is popular among the data visualisation community for its ability to
handle localised distortions \cite{brown92survey}. Given sample points on a planar surface, the domain must be divided
into polygonal segments with sample domain values as vertices. A smooth surface segment is then fitted
over each polygon such that the sample points associated with the vertices lie on the surface, and the
edges associated with adjacent segments meet in a prescribed way. There is usually more than one way
to segment the domain space, and different segmentations may yield very different interpolations.

Global (as opposed to piecewise) polynomial interpolation is a theoretical possibility that
has proven elusive in practice. The Stone-Weierstrass Theorem \cite{cheney1966,pinkus2005} establishes the existence of a
polynomial approximation for every real-valued continuous function defined on a closed interval. Therefore, as long as the sample
data does not admit the existence of discontinuities, we should be able to fit a polynomial to the points
with an arbitrary degree of precision. The problem (exemplified by Runge's function \cite{cheney1966,forsythe1977}) is that
successive interpolations do not necessarily converge as more data points are added to the set of sample
points (c.f. Faber's Theorem \cite{cheney1966,forsythe1977}). We will refer to this problem later as the \emph{convergence problem}.

Our solution is similar to global polynomial interpolation. However, the fact that we are
dealing with noisy data allows us to relax the requirement for an exact fit and to focus
instead on the suggested shape of the function. It is because of this relaxed focus on an exact fit that we
refer to our problem as an approximation problem rather than an interpolation problem. By focusing on
the suggested shape of the function, we directly address the convergence problem.

In this paper, we present the generic function approximation algorithm and use image rectification as an example of its use. In section \ref{sect:theory}, we develop the theory behind the approximation algorithm in the univariate
case. Section \ref{sect:bivariate} discusses the bivariate version of the approximation algorithm that is used to solve our
robot vision problem. Section \ref{sect:summary} summarizes our findings.

\section{Polynomial approximation of functions}\label{sect:theory}

\noindent In the univariate case, the polynomial approximation problem can be defined as follows:

\begin{definition}[The univariate polynomial approximation problem]\label{defn:upa}
Find a polynomial $\mathrm{P}(x)$ that fits a set
of $m$ sample points $(x_i,y_i)$; where $i=1,2,\ldots,m$ and the $x_i$ are distinct; such that:
\begin{displaymath}
\max_{i=1}^m|y_i-\mathrm{P}(x_i)| \le \epsilon, \mathrm{ for\ } \epsilon\ge 0
\end{displaymath}
\end{definition}
That is, we are trying to identify a polynomial, $\mathrm{P}(x)$, that fits a finite set of points to a degree of precision
determined by $\epsilon$. The condition that the $x_i$ must be distinct ensures that there are no discontinuities in the target function. Otherwise, a solution is not assured.

The Weierstrass approximation theorem \cite{lagaris1997,cheney1966,forsythe1977} assures us that this is a solvable problem
for $\epsilon > 0$:
\begin{theorem}[Weierstrass]
If $\mathrm{F}$ is any continuous function on the finite closed interval $[a,b]$, then for every $\epsilon > 0$ there exists a
polynomial $\mathrm{P}_n(x)$ of degree $n$ (where $n$ depends on $\epsilon$) such that:
\begin{displaymath}
\max_{x\in[a,b]}|\mathrm{F}(x)-\mathrm{P}_n(x)|<\epsilon
\end{displaymath}
\end{theorem}

Our finite set of points is not generated from a known function that we are trying to approximate, so we
will be looking instead for an approximation that generates intermediate points that are ``good'' in some
generic sense. We require the algorithm to yield an approximating polynomial that does not ``curve''
unnecessarily between sample points.
In our algorithm, we try to conservatively fit a shape to the sample points using a linear
combination of Chebyshev polynomials. A preference for fitting lower order Chebyshev polynomials reflects a preference for simple shapes. This emphasis on simple shapes addresses the convergence problem by only adding detail when it results in a better fit. 

Section \ref{sect:cvb} presents the basis for our algorithm in its purest form and explains why it is
inadequate. We refer to this algorithm as the Cartesian vector based (CVB) interpolation. This algorithm is of
theoretical interest only, since it produces results similar to Lagrange interpolation, with the same
convergence problems illustrated by Runge and covered by Faber's theorem [5,7]. It sets the scene for
presentation of the modified version of the algorithm.
In section \ref{sect:CVBaa}, we present the modified algorithm that gives better results on
the same data in terms of capturing the suggested shape of the curve. We refer to this second algorithm as the CVB approximation algorithm.

\subsection{Casting the problem in Cartesian vector space}\label{sect:cvb}
\noindent We address the approximation problem by finding a linear combination of the first $n$ Chebyshev
polynomials of the first kind that exhibits the desired properties of $\mathrm{P}(x)$. Thus:
\begin{equation}\label{eq:P}
\mathrm{P}(x)=\sum_{i=0}^{n-1}a_i\mathrm{T}_i(x) 
\end{equation}
where $\mathrm{T}_i(x)$ is defined for $x\in[-1,1]$ as
\begin{itemize}
\item $\mathrm{T}_0(x)=1$,
\item $\mathrm{T}_1(x)=x$,
\item $\mathrm{T}_i(x)=2x\mathrm{T}_{i-1}(x)-\mathrm{T}_{i-2}(x)$, for $i>1$.
\end{itemize}

The problem is to find a set of coefficients, $a_i$, $i=0,\ldots,n-1$ that establish a fit. Although other sets of basis polynomials exist, Chebyshev polynomials reputedly yield good interpolation results.  In what follows, we will assume without loss of generality that $x$ falls within the interval $[-1,1]$.

The fact that we are using a finite set of sample points allows us to reformulate the problem as one involving vectors in a Cartesian
$m$-space; where $m$ is the number of sample points. For this purpose the problem is restated as follows:
\begin{definition}[The univariate Cartesian vector based (CVB) polynomial approximation problem]\label{defn:cvbupa}
	Given 
	\begin{itemize}
	\item a set of points, $(x_i,y_i)\in \mathbb{R}^2$, $i = 1,\ldots,m$;
	\item a set of $m$-dimensional Cartesian vectors, $\mathbf{\tau}_j$, $j=0,\ldots,n-1$, such that the $i$'th component of $\mathbf{\tau}_j$ is equal to $\mathrm{T}_j(x_i)$, for $i=1,\ldots,m$;
	\item a Cartesian vector, $\mathbf{\gamma}$ such that the $i$'th component of $\mathbf{\gamma}$ is equal to $y_i$, for $i=1,\ldots,m$;
	\item a real value $\epsilon$ ($\epsilon>0$);
	\end{itemize}
	find values for a set of scalar quantities, $a_j$, such that:
	\begin{itemize}
	\item $\mathbf{\rho}=\sum_{j=0}^{n-1}a_j\mathbf{\tau}_j$;
	\item $\mathbf{\delta}=\mathbf{\gamma} - \mathbf{\rho}$; 
	\item $|\delta^i|\le \epsilon$ for each component, $\delta^i$, of $\mathbf{\delta}$ ($i=1,\ldots,m$).
	\end{itemize}
\end{definition}

For example, the representation for the expression
\begin{displaymath}
x^2+2x+1
\end{displaymath}
as a linear combination of Chebyshev polynomials is:
\begin{displaymath}
1.5\mathrm{T}_0(x)+2\mathrm{T}_1(x)+0.5\mathrm{T}_2(x)
\end{displaymath}

\begin{table}[!tb]
\caption{\small The CVB representation of a curve fit.}\label{table:cvbexample}
\small
\begin{tabular}{|c|r @{.} l | r @{.} l|r @{.} l|r @{.} l|r @{.} l|r @{.} l|r @{.} l|r @{.} l|r @{.} l|}
\hline
 i  &  \multicolumn{2}{|c|}{$x_i$}  &  \multicolumn{2}{|c|}{$y_ i$} &   \multicolumn{2}{|c|}{$\mathrm{T}_0(x_i)$}  &  \multicolumn{2}{|c|}{$\mathrm{T}_1(x_i)$}  &  \multicolumn{2}{|c|}{$\mathrm{T}_2(x_i)$}  &  \multicolumn{2}{|c|}{$a_0\mathrm{T}_0(x_i)$} & \multicolumn{2}{|c|}{$a_1\mathrm{T}_1(x_i)$} &  \multicolumn{2}{|c|}{$a_2\mathrm{T}_2(x_i)$} & \multicolumn{2}{|c|}{$\mathrm{P}(x_i)$}\\
  \hline
 1 & -1&0 & 0&0 & 1&0 & -1&0 & 1&0 & 1&5 & -2&0 & 0&5 & 0&0\\
 2 & 0&0 & 1&0 & 1&0 & 0&0 & -1&0 & 1&5 & 0&0 & -0&5 & 1&0\\
 3 & 1&0 & 4&0 & 1&0 & 1&0 & 1&0 & 1&5 & 2&0 & 0&5 & 4&0\\
 \hline
\end{tabular}
\end{table}

Table \ref{table:cvbexample} shows the situation for this function for three evenly spaced sample points. In this example, each
Chebyshev polynomial yields a 3-dimensional Cartesian vector. A linear combination of these vectors
yields a solution. We will refer to the vectors generated by the polynomial terms as \emph{term vectors}.

If the term vectors $\mathbf{\tau}_j$ are orthogonal, they may be taken as basis vectors and solution values for the coefficients, $a_j$, may be directly obtained from the projection of $\mathbf{\gamma}$ on each $\mathbf{\tau}_j$, respectively. 

However, orthogonality does not hold in general, as illustrated by the set of points in Table \ref{table:cvbexample}. An obvious solution to this dilemma is to first identify an orthogonal set of vectors corresponding to the term vectors and use this set to determine a solution. We developed an algorithm to do just this. We refer to it as the \emph{CVB interpolation}.

\subsection{The CVB interpolation}\label{sect:ocfa}

\noindent This algorithm computes an orthogonal set of vectors, $\{ \mathbf{o}_0, \mathbf{o}_1,\ldots, \mathbf{o}_{m -1}\}$, from the first $m$ linearly independent term vectors, derives the projection of $\mathbf{\gamma}$ on each of these vectors and modifies the coefficients of the term vectors accordingly. 

The orthogonalisation of the term vectors is based on the following theorem on which the well-known \emph{Gram-Schmidt orthogonalisation process} is based:
\begin{theorem}
	Let $\{\mathbf{v}_1,\mathbf{v}_2,\ldots,\mathbf{v}_n\}$ be an orthonormal set of vectors in a vector space, $V$. Then for any vector $\mathbf{w}\in V$, the vector:
	\begin{displaymath}
		\mathbf{o}= \mathbf{w} - \sum_{i=1}^n(\mathbf{w}\cdot\mathbf{v}_i)\mathbf{v}_i
	\end{displaymath}
	is orthogonal to each $\mathbf{v}_i$, $i=1,\ldots,n$.
\end{theorem}
This theorem can easily be generalised to apply to orthogonal (as distinct from \emph{orthonormal}) sets as follows:
\begin{theorem}
	Let $\{\mathbf{v}_1,\mathbf{v}_2,\ldots,\mathbf{v}_n\}$ be an orthogonal set of vectors in a vector space, $V$. Then for any vector $\mathbf{w}\in V$, the vector:
	\begin{displaymath}
		\mathbf{o}= \mathbf{w} - \sum_{i=1}^n\frac{(\mathbf{w}\cdot\mathbf{v}_i)}{(\mathbf{v}_i\cdot\mathbf{v}_i)}\mathbf{v}_i
	\end{displaymath}
	is orthogonal to each $\mathbf{v}_i$, $i=1,\ldots,n$.
\end{theorem}
The additional factor in the summation normalises each $\mathbf{v}_i$.

In our algorithm, we identify an orthogonal set of vectors, $\{\mathbf{o}_0,\mathbf{o}_1,\ldots,\mathbf{o}_{n-1}\}$ and a set of scalars $p_{j,k}$, such that:
\begin{eqnarray}
\mathbf{o}_0 & = & \mathbf{\tau}_0 \label{eq:basecase}\\
p_{j,k} & = & \frac{(\mathbf{\tau}_j\cdot\mathbf{o}_k)}{(\mathbf{o}_k\cdot\mathbf{o}_k)} \label{eq:pjk}\\
\mathbf{o}_j & = & \mathbf{\tau}_j-\sum_{k=0}^{j-1}p_{j,k}\mathbf{o}_k, \mathrm{for\ } j=1,\ldots,n-1 \label{eq:recursivestep}
\end{eqnarray}
We refer to each $o_j$ as an \emph{orthogonal component} of the corresponding $\mathbf{\tau}_j$, for $j=0,\ldots,n-1$.

For the CVB interpolation, we need to express $\mathbf{o}_j$ as a function of $\mathbf{\tau}_k$.
Using (\ref{eq:recursivestep})  we write:
\begin{eqnarray}
\mathbf{o}_j  & = & \sum_{k=0}^{j}q_{j,k}\mathbf{\tau}_k\nonumber\\
& =  & \mathbf{\tau}_j-\sum_{k=0}^{j-1}p_{j,k}\mathbf{o}_k \nonumber\\
 & =  & \mathbf{\tau}_j-\sum_{k=0}^{j-1}p_{j,k}\sum_{l=0}^kq_{k,l}\mathbf{\tau}_l \nonumber\\
 & = & \mathbf{\tau}_j-\sum_{l=0}^{j-1}\mathbf{\tau}_l\sum_{k=l}^{j-1}p_{j,k}q_{k,l} \nonumber\\
 & = & \mathbf{\tau}_j-\sum_{k=0}^{j-1}\mathbf{\tau}_k\sum_{l=k}^{j-1}p_{j,l}q_{l,k} \label{eq:oj}
\end{eqnarray}
From (\ref{eq:oj}) we get:
\begin{eqnarray}
q_{j,j} & = & 1\\
q_{j,k} & = & -\sum_{l=k}^{j-1}p_{j,l}q_{l,k}
\end{eqnarray}

\begin{figure}[!tb]
\framebox{
\begin{minipage}{\textwidth}
\texttt{\small
\begin{enumerate}
\item Set all the coefficients, $a_j$, to zero.
\item Compute the orthogonal components, $o_j$ and the scalars $q_{j,k}$.
\item For $j = 0,\ldots,n-1$ do
  \begin{enumerate}
  \item Compute the error vector, $\mathbf{\delta}$.
  \item Set {\sf Oinc}  to $\frac{\mathbf{o}_j\cdot\mathbf{\delta}}{\mathbf{o}_j\cdot\mathbf{o}_j}$ 
  \item 
        \begin{tabbing}
        For \=$k= 0,\ldots,j$ do\\
        \>Set $a_k$ to $a_k + {\sf Oinc}\cdot q_{j,k}$.\\
        Endfor	
        \end{tabbing}
  \end{enumerate}
  Endfor
\item Return the set of coefficients, $a_j$.
\end{enumerate}
}
\end{minipage}
}
\caption{\small The CVB interpolation (see section \ref{sect:ocfa}). Computational details have been omitted that deal with avoidance of representational and computational error.}\label{fig:ocfa}
\end{figure}

Fig. \ref{fig:ocfa} shows the CVB interpolation in pseudocode. Table \ref{table:ocfaexample} shows a trace of the CVB interpolation on the problem depicted in Table \ref{table:cvbexample}.
 
\begin{table}[!tb]
\begin{center}
\caption{\small A trace of the CVB interpolation on the problem depicted in Table \ref{table:cvbexample}.}\label{table:ocfaexample}
\small
\begin{tabular}{|r @{.} l | r @{.} l|r @{.} l|r @{.} l|r @{.} l|r @{.} l|r @{.} l|}
\hline
 \multicolumn{2}{|c|}{$a_0$} &  \multicolumn{2}{|c|}{$a_1$}  &  \multicolumn{2}{|c|}{$a_2$} &   \multicolumn{2}{|c|}{$\|\mathbf{\delta}\|$}  &  \multicolumn{2}{|c|}{$\mathrm{P}(x_1)$}  &  \multicolumn{2}{|c|}{$\mathrm{P}(x_2)$}  &  \multicolumn{2}{|c|}{$\mathrm{P}(x_3)$} \\
  \hline
  1&666666667 & 0&0 & 0&0 & 2&943920289 & 1&666666667 & 1&666666667 & 1&666666667\\
  1&666666667 & 2&0 & 0&0 & 0&816496581 & -0&333333333 & 1&666666667 & 3&666666667\\
  1&5 & 2&0 & 0&5 & 0&0 & 0&0 & 1&0 & 4&0\\
  \hline
\end{tabular}
\end{center}
\end{table}

 \subsection{Difficulties with polynomial interpolation}
 
\begin{figure}[!htb]
 	\begin{tabular}{cc}
 	\includegraphics[width = 0.5\textwidth]{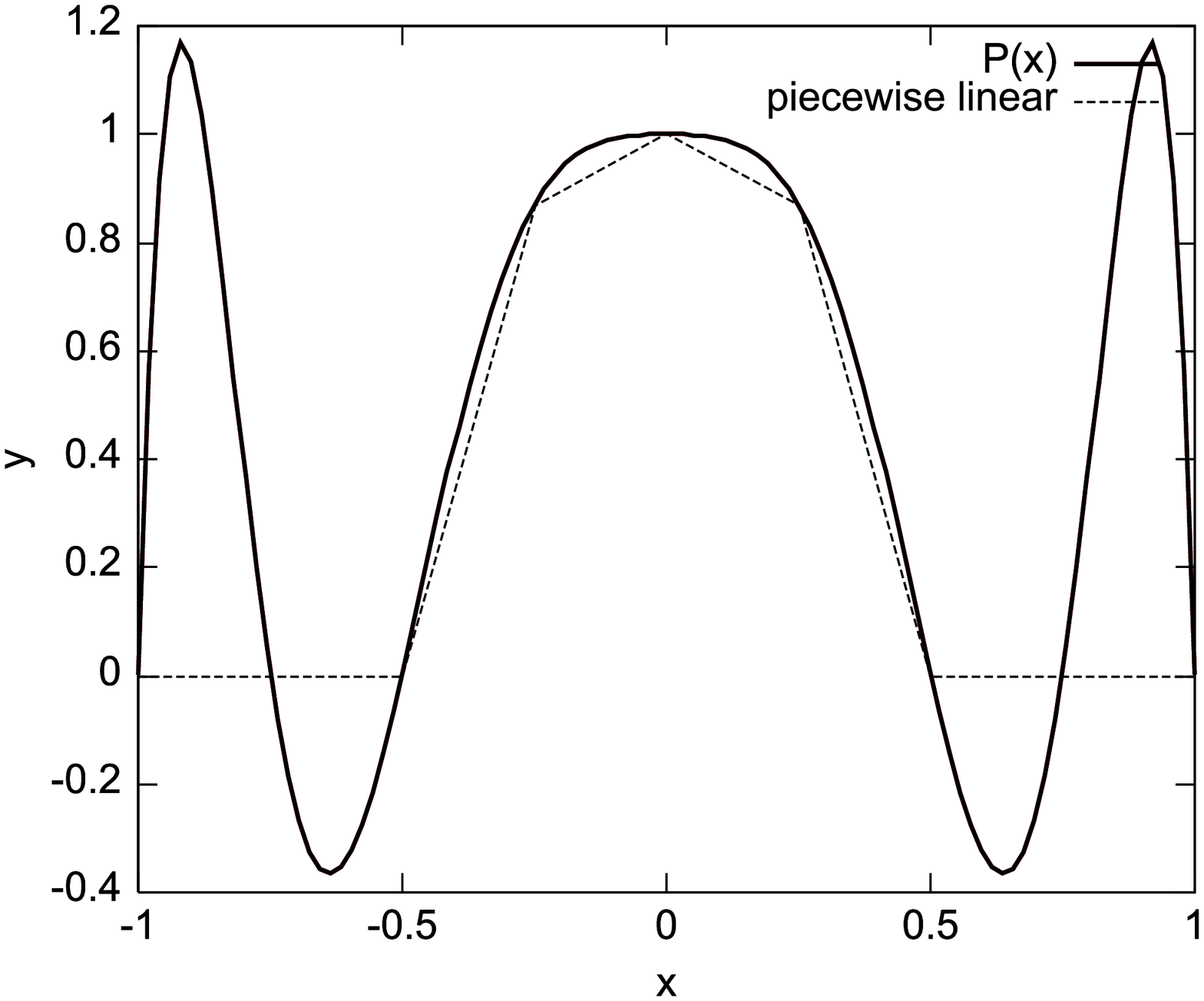} & \includegraphics[width = 0.5\textwidth]{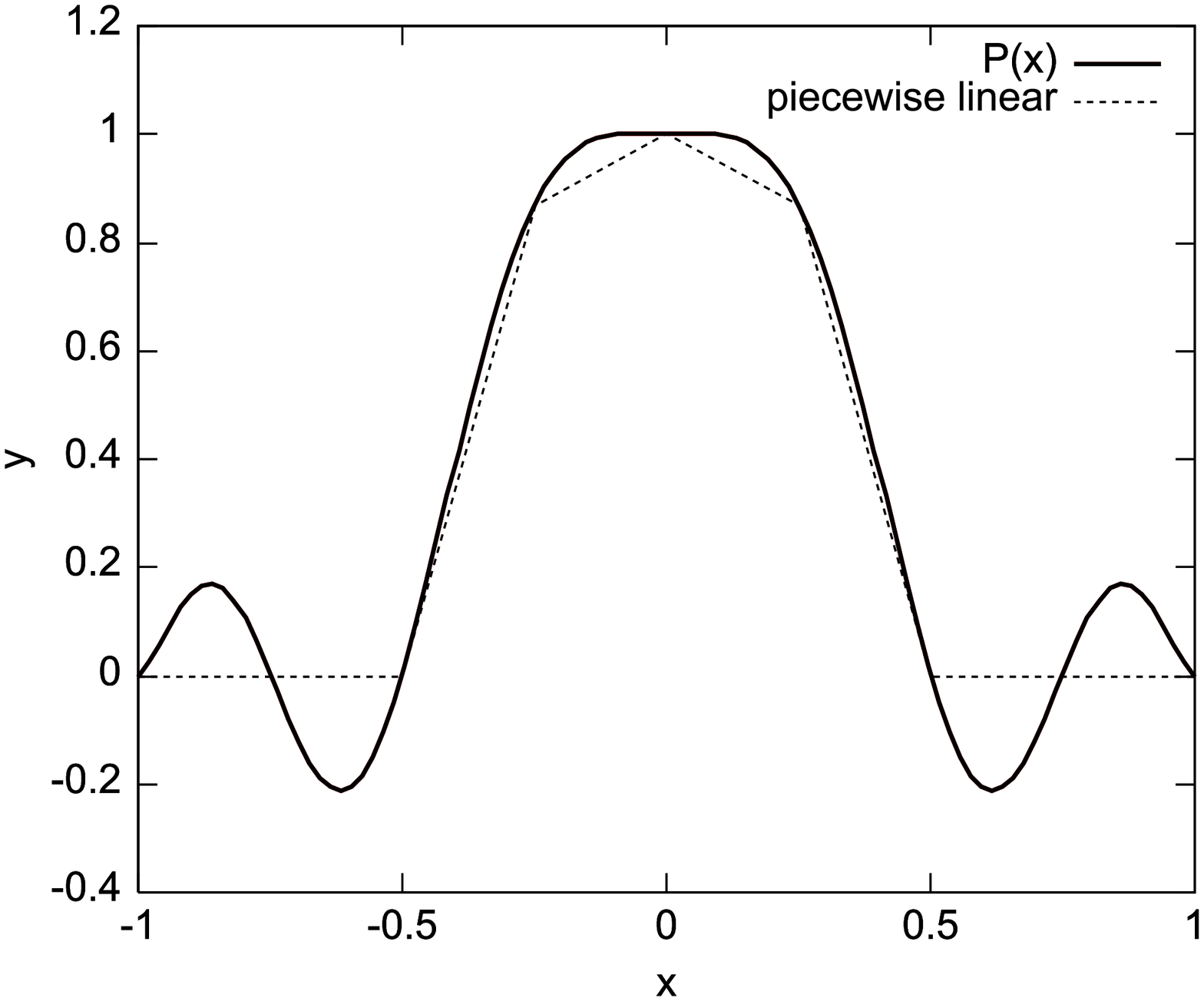}\\
 	{\small (a) interpolation on 9 points} & {\small (b) approximation on data from (a)}
  	\end{tabular}
\caption{\small Interpolation and approximation of data that is flat in some places and humped in others.}\label{fig:humpnflat}
 \end{figure}

\begin{figure}[!htb]
 	\begin{tabular}{cc}
 	\includegraphics[width = 0.5\textwidth]{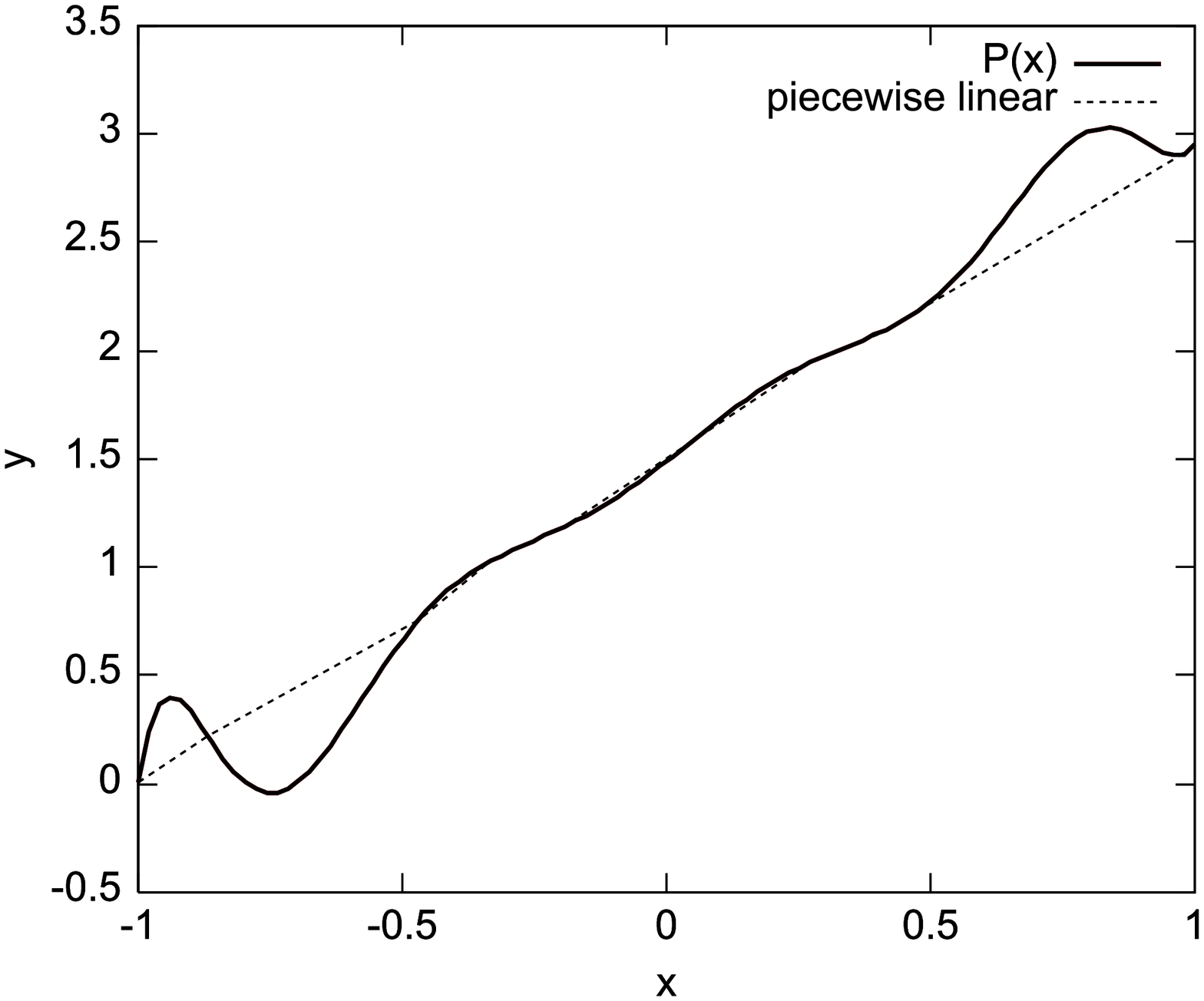}&\includegraphics[width = 0.5\textwidth]{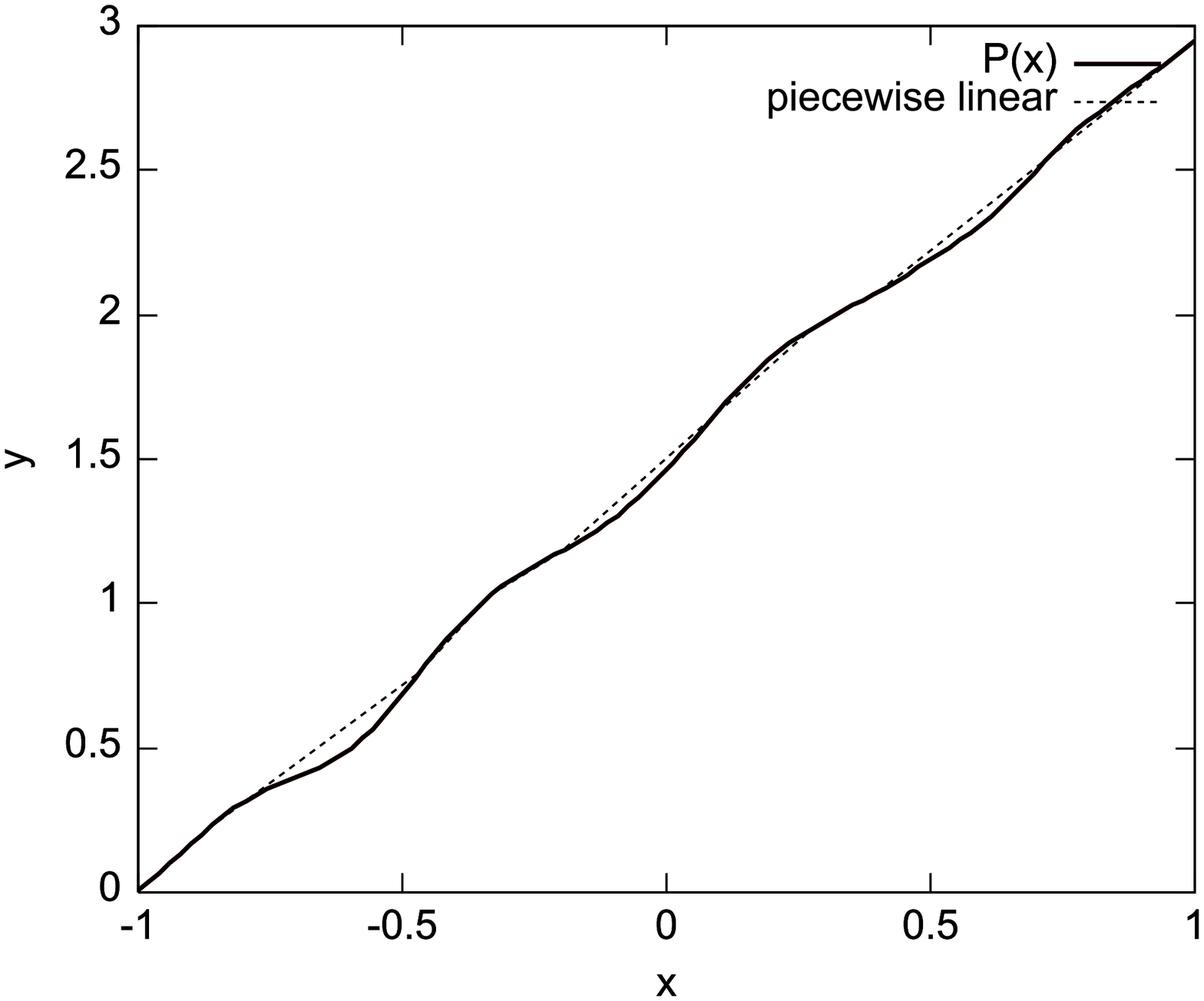}\\
 	{\small (a) interpolation on 9 points} & {\small (b) approximation on data from (a)}
  	\end{tabular}
\caption{\small Interpolation and approximation of noisy straight line data with $y$ values given at unevenly spaced $x$ values.}\label{fig:noisyline}
 \end{figure}

\begin{figure}[!htb]
	\begin{tabular}{cc}
 	\includegraphics[width = 0.5\textwidth]{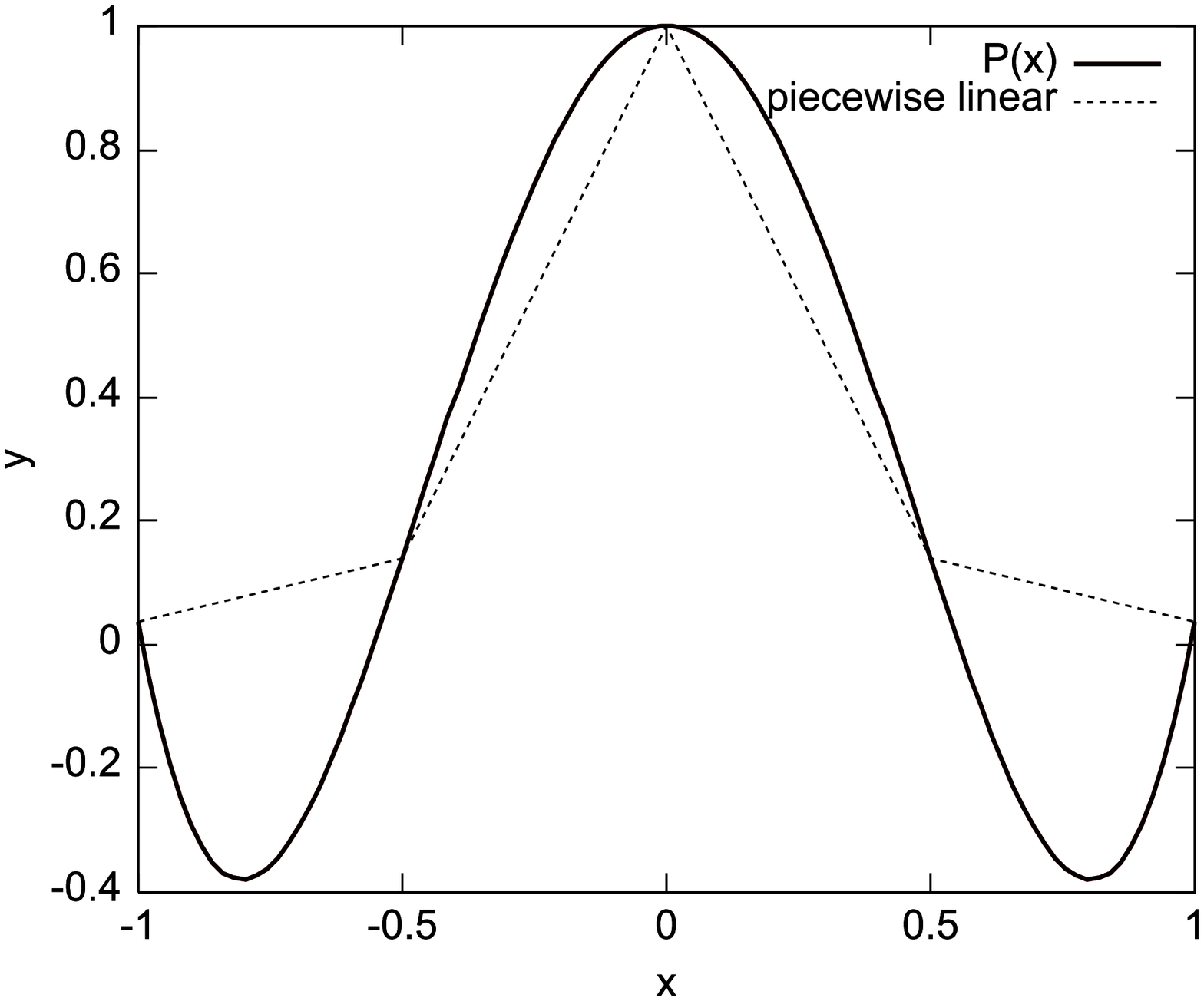} &  \includegraphics[width = 0.5\textwidth]{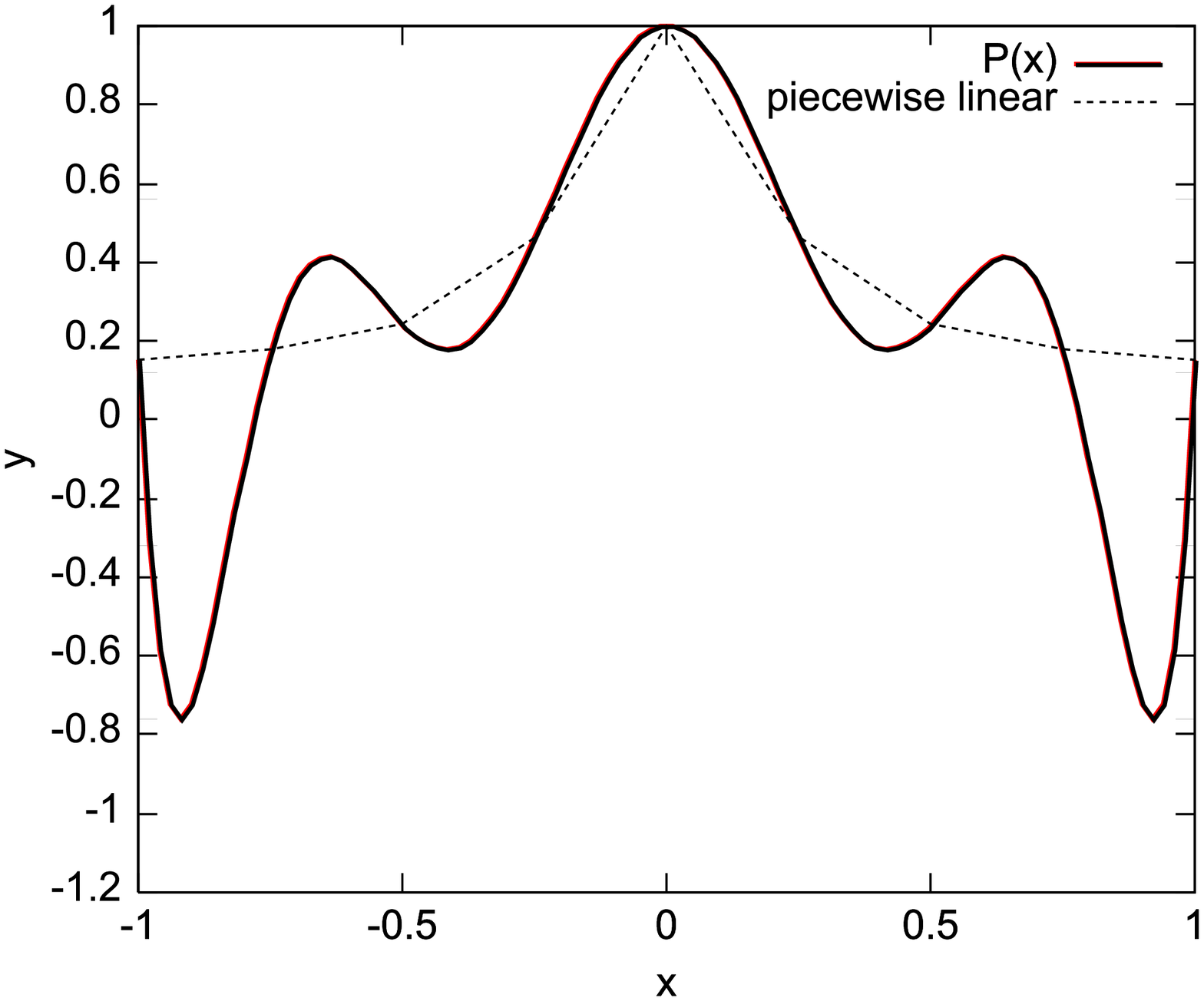}\\
 	{\small (a) interpolation on 5 points} & {\small (b) interpolation on 9 points} \\
	\includegraphics[width = 0.5\textwidth]{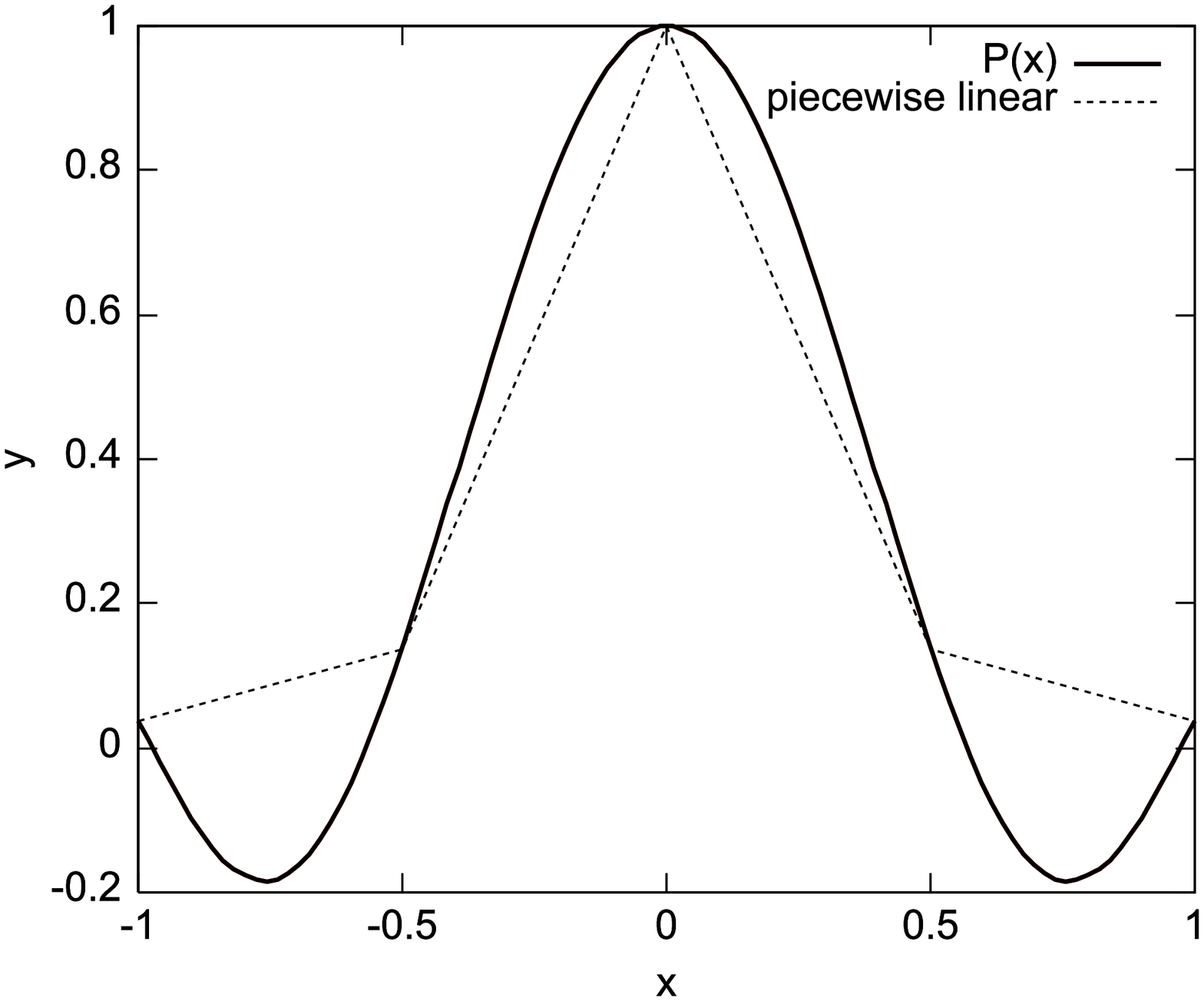} &  \includegraphics[width = 0.5\textwidth]{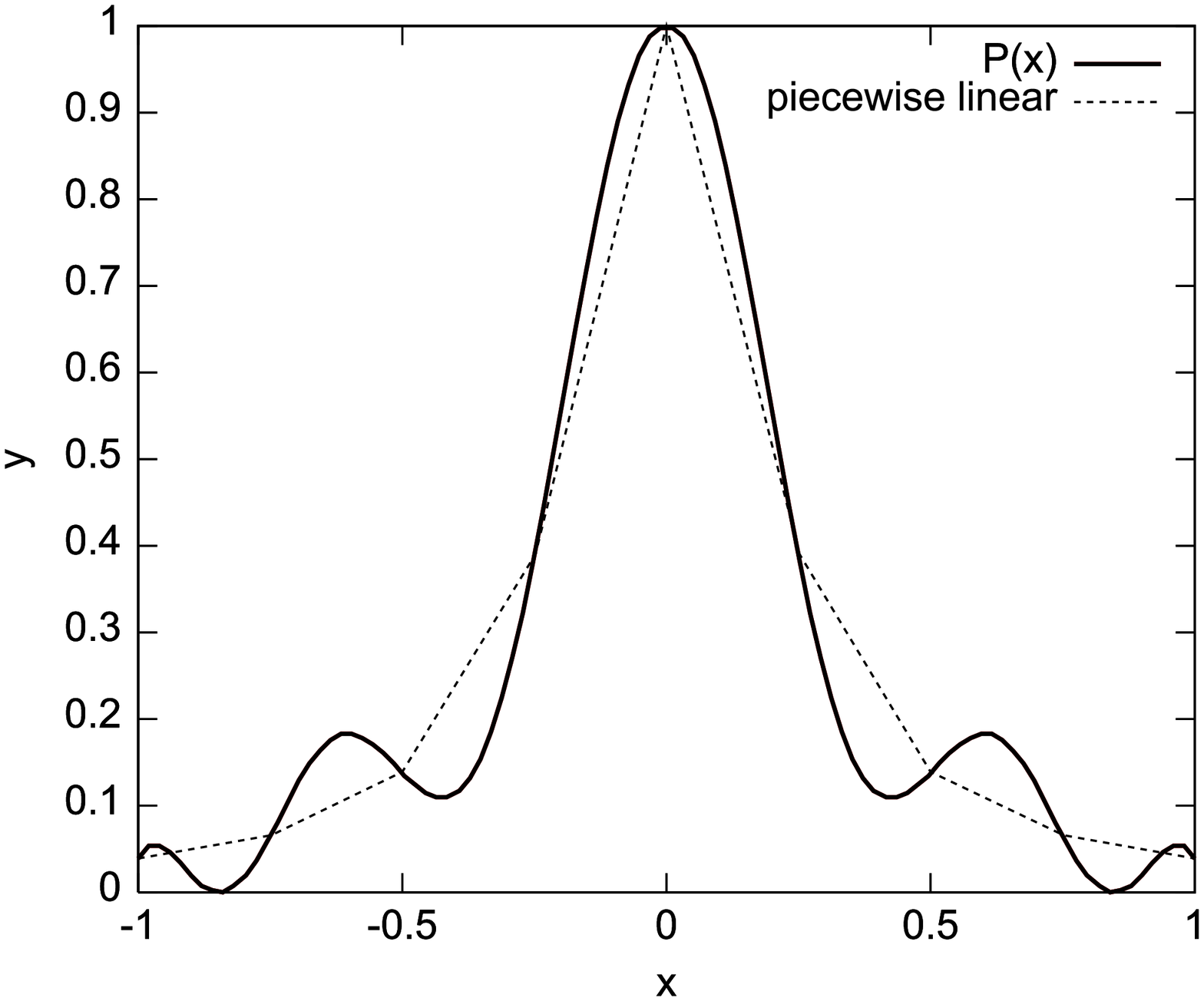}\\
 	{\small (c) approximation on data from (a)}& {\small (d) approximation on data from (b)}
 	\end{tabular}
 \caption{\small Performance on the convergence problem. Data sets were taken from the Runge function. Comparison of (a) with (b) illustrates that interpolating with more data points does not always give better results. Comparison of (c) and (d) illustrate convergence on the same data using the CVB approximation algorithm presented in section \ref{sect:CVBaa}. }\label{fig:noconvergence}
 \end{figure}

\noindent The following types of data serve to illustrate the difficulties encountered with global polynomial interpolation \cite{fausett2003}:
 \begin{itemize}
 \item ``Humped and flat data'' that suggest a flat curve in some regions and not in others (Fig. \ref{fig:humpnflat}). 
 \item ``Noisy straight line data'' with $y$ values given at unevenly spaced $x$ values (Fig. \ref{fig:noisyline}). 
 \item Data from functions that exhibit the convergence problem, the classical example of which is the Runge function (Fig. \ref{fig:noconvergence}). Faber's theorem establishes that the Runge function is not the only function that exhibits this problem.
 \end{itemize}

 These types of data all exhibit a common problem: there is significant deviation of the fitted polynomial from the shape of the curve suggested by the data points. With regard to the convergence problem, progressively adding more data points does not lead to progressively better agreement between the fitted polynomial and the suggested shape of the curve (c.f. Faber's Theorem). The problem of large deviations from ground truth at the intermediate points is usually addressed in one of two ways: either additional information is supplied as data, or additional constraints are imposed as part of the method of solution. 
 
 Hermite interpolation is an attempt to improve the quality of the interpolation by using additional information (first derivatives at the data points), but suffers from the fact that the additional information is local to the data points and so does not sufficiently constrain what happens at the intermediate points. What seems to be needed are global constraints that affect the quality of the solution over the entire interval. 
 
 The difficulty with applying global constraints in interpolation is the requirement that a perfect fit be achieved at the sample points. This requirement gives the sample points a special status relative to the intermediate points. The fact that a constraint must respect this special status would render it less than global, as the constraint would have to behave differently in the vicinity of the sample points. If a perfect fit is not required, then global \emph{approximation} may be used. Theory states that polynomial approximations exist that follow closely any curve that can be described by a continuous real-valued function on a closed interval. Global constraints can be more easily applied to selecting the best approximation than to finding a perfect fit.

\subsection{The CVB approximating algorithm}\label{sect:CVBaa}

\noindent In this section, we present our \emph{ CVB approximating algorithm}. This algorithm gives better convergence than the
CVB interpolation. This improvement was achieved by exploiting the observation that the shape of a function that fits the error vector, $\mathbf{\delta}$ (see Definition \ref{defn:cvbupa}), is captured in part by the direction of $\mathbf{\delta}$,
together with the fact that the direction of a Cartesian vector is invariant under scalar multiplication. We define the shape of a function as follows:

\begin{definition}[The shape of a function]\label{def:shape}
	Consider two continuous real-valued functions, $\mathrm{F}_1$ and $\mathrm{F}_2$, defined on the same closed interval, $I$. $\mathrm{F}_1$ and $\mathrm{F}_2$ are of the same shape if there exists a finite non-zero real scale factor, $s$, such that $\mathrm{F}_1 = s\mathrm{F}_2$ 
\end{definition}
Thus, a particular shape is denoted by the set of all the functions that are of that shape. We will only explore in this paper the properties of this concept of shape that are relevant to our exposition. In the first instance, we will confine our discussion to real-valued functions defined on the interval $I=[-1,1]$.

\begin{definition}[Value ratios]\label{prop:valueratios}
If two functions, $\mathrm{F}_1$ and $\mathrm{F}_2$, are of the same shape, then for any two distinct values, $x_1,x_2\in I$, such that $\mathrm{F}_1(x_2)\ne 0$ ( and, by extention, $\mathrm{F}_2(x_2)\ne 0$), it is the case that $\frac{\mathrm{F}_1(x_1)}{\mathrm{F}_1(x_2)} = \frac{\mathrm{F}_2(x_1)}{\mathrm{F}_2(x_2)}$. That is, the ratio of $\mathrm{F}_1(x_1)$ to $\mathrm{F}_1(x_2)$ is the same as the ratio of $\mathrm{F}_2(x_1)$ to $\mathrm{F}_2(x_2)$.
\end{definition}
It is this invariance of value ratios across functions of the same shape that defines our concept of shape.

Now consider a Chebyshev polynomial, $\mathrm{T}$, and its associated term vector, $\mathbf{\tau}$, defined on a set of sample points. The ratios of the components of $\mathbf{\tau}$ to one another are value ratios of $\mathrm{T}$. But these ratios also define the direction of the Cartesian vector, $\mathbf{\tau}$. So that the shape of $\mathrm{T}$ determines the direction of $\mathbf{\tau}$. However, since the finite dimensional term vector $\mathbf{\tau}$ cannot capture the infinity of value ratios that specify the shape of $\mathrm{T}$, the direction of $\mathbf{\tau}$ only partly captures the shape of $\mathrm{T}$.

If we wish to compare the shape of a Chebyshev polynomial, $\mathrm{T}$, with the shape suggested by
$\mathbf{\delta}$, we can compare the direction of the corresponding term vector, $\mathbf{\tau}$, with the direction of $\mathbf{\delta}$. In particular, we look at the projection of $\mathbf{\delta}$ on $\mathbf{\tau}$ and express the projection as a scalar multiple of $\mathbf{\tau}$. The scalar factor thus identified is used to adjust the coefficient of $\mathrm{T}$ in equation (\ref{eq:P}). 

The main difference between the CVB approximating algorithm and the CVB interpolation is an emphasis on fitting the shape of the term vectors as distinct from an orthogonal component thereof. However, since term vectors are not orthogonal in general, more than one approximation may be possible. As a means of selecting a preferred solution, we adopt an heuristic based on fitting lower order terms first.

\begin{figure}[!tb]
\framebox{
	\begin{minipage}{\textwidth}
	\texttt{\small
	\begin{enumerate}
	\item Set all the coefficients, $a_j$, to zero.
	\item Compute the error vector, $\mathbf{\delta}$.
	\item Set $j$ to $0$.
	\item While $j<n$ and $\max(|\delta^i|)>\epsilon$ for $i=1,\cdots,m$ do
	  \begin{enumerate}
	  \item Set $a_j$ to $a_j + \frac{\mathbf{\tau}_j\cdot\mathbf{\delta}}{\mathbf{\tau}_j\cdot\mathbf{\tau}_j}$.\\
	  \item Compute the error vector, $\mathbf{\delta}$.
	  \item 
		\begin{tabbing}
		For \=$k= (j-1),(j-2),\ldots,0$ do\\
		\>\=Compute the error vector, $\mathbf{\delta}$.\\
		\>Set $a_k$ to $a_k + \frac{\mathbf{\tau}_k\cdot\mathbf{\delta}}{\mathbf{\tau}_k\cdot\mathbf{\tau}_k}$.\\
		\>Compute the error vector, $\mathbf{\delta}$.\\
		Endfor	
		\end{tabbing}
	  \item Set $j$ to $j+1$
	  \end{enumerate}
	  EndWhile
	\item Return the set of coefficients, $a_j$.
	\end{enumerate}
	}
	\end{minipage}
}
\caption{\small The CVB approximation algorithm (see section \ref{sect:CVBaa}). Computational details have been omitted that deal with avoidance of representational and computational error.}\label{fig:CVBaa}
\end{figure}

Like the CVB interpolation, the CVB approximation algorithm works by finding projections of the error vector on a vector space identified by a finite number of term vectors. It gives preference to using the term vectors associated with lower order Chebyshev terms. The process is as follows (see Fig. \ref{fig:CVBaa} for the algorithm in pseudocode):
\begin{itemize}
\item The first projection to be eliminated is the projection of $\mathbf{\delta}$ on $\mathbf{\tau}_0$. 
Whenever the projection of $\mathbf{\delta}$ on
a term vector, $\mathbf{\tau}$, has been eliminated, $\mathbf{\tau}$ will be referred to as having been \emph{visited}.
\item If $\mathbf{\tau}_j$ has been visited, then $\mathbf{\tau}_{j-1}$ must be re-visited. This rule is applied recursively to revisits until $\mathbf{\tau}_0$ is revisited.
\item After $\mathbf{\tau}_j$ has been visited and all consequent revisits have taken place, then $\mathbf{\tau}_{j+1}$ is visited.
\item The process terminates when the required degree of accuracy of fit has been obtained.
\end{itemize}
Revisits are necessary in order to maintain the fit of lower order terms. More preferred terms are revisited after lesser preferred ones in order to minimise the effect of revisits on their fit. Thus the most preferred term is revisited last.

\subsubsection{Convergence of the CVB approximation algorithm}\label{sect:convergence}

\noindent To prove convergence, we consider the general situation in which we have visited the first $p$ term
vectors and still have a non-zero residual error vector. We note that we can eliminate this error if we can
approximate the error vector using a linear combination of the Chebyshev polynomials that have not yet
been visited. We use the real-valued version of the Stone-Weierstrass theorem \cite{pinkus2005} to prove that this is
possible.

\begin{theorem}[]
Let $X$ be a compact set and let $\mathrm{C}(X)$ denote the space of continuous real-valued functions defined on
$X$. Assume that $A$ is a subalgebra of $\mathrm{C}(X)$. Then $A$ is dense in $\mathrm{C}(X)$ in the uniform norm iff $A$
separates points and for each $x\in X$ there exists an $\mathrm{f}\in A$ satisfying $\mathrm{f}(x)\ne 0$.
\end{theorem}

\noindent\textbf{Proof}\\
To prove that the polynomial vector space $P$ with basis set $\{\mathrm{T}_n\mid n>p\}$ is
dense in $\mathrm{C}(X)$, where X is the interval [-1,1], prove that:
\begin{enumerate}
\item $X$ is compact.
\item $P$ is a subalgebra of $\mathrm{C}(X)$.
\item $P$ separates points (for any distinct points, $x_1,x_2\in X$ , there is a $\mathrm{T}_n, n>p$ s.t.
$\mathrm{T}_n(x_1)\ne \mathrm{T}_n(x_2)$.
\item For each $x\in X$ there is a $\mathrm{T}_n, n>p$ s.t. $\mathrm{T}_n(x)\ne 0$. 
\end{enumerate}

Taking each condition in turn:
\begin{enumerate}
\item $X$ is a closed and bounded interval on the real number line and is therefore compact.
\item $P$ is a subalgebra of $\mathrm{C}(X)$ since $P$ is closed for the usual multiplication and addition of
functions, and for scalar multiplication by real values. This is sufficient to ensure that the
defining properties of an algebra hold for the subspace, $P$ as they do for $\mathrm{C}(X)$.
\item Let $x_1,x_2\in[-1,1]$ such that $x_1\ne x_2$ and $\mathrm{T}_n(x_1)=\mathrm{T}_n(x_2)$ . Given that $\mathrm{T}_n(x)=\cos(n\cos^{-1}(x))$,
let $x_1=\cos(\theta_1)$ and $x_2=\cos(\theta_2)$ with $\theta_1>\theta_2$ and $\theta_1,\theta_2\in[0,\pi]$. If $\mathrm{T}_n(x_1)=\mathrm{T}_n(x_2)$ , we get $\cos(n\theta_1)=\cos(n\theta_2)$. This implies that either $n\theta_1 \pmod{2\pi} = n\theta_2 \pmod{2\pi}$ or $-(n\theta_1\pmod{2\pi} = 2\pi-(n\theta_2\pmod{2\pi})$, which gives
\begin{equation}\label{eq:thetadiff}
\theta_1-\theta_2 = \frac{2\pi N}{n} \textrm{ for some integer, } N
\end{equation}
Since $\theta_1,\theta_2\in[0,\pi]$ and $\theta_1>\theta_2$, it follows from (\ref{eq:thetadiff}) that:
\begin{equation}\label{eq:Nnrange}
0<\frac{N}{n}<\frac{1}{2}
\end{equation}
Now, $\mathrm{T}_{n+1}(x_1)=\cos((n+1)\theta_1)$. So that from (\ref{eq:thetadiff}):
\begin{equation}\label{eq:np1theta}
(n+1)\theta_1 = 2\pi\left(N+\frac{N}{n}\right)+(n+1)\theta_2
\end{equation}
From (\ref{eq:Nnrange}) and (\ref{eq:np1theta}) we can conclude that $\cos((n+1)\theta_1)\ne\cos((n+1)\theta_2)$. So that $\mathrm{T}_{n+1}$ separates the points.
\item The zeroes of $\mathrm{T}_n$ are given by $\zeta_j^{(n)}=\cos\left(\frac{(2j-1)\pi}{2n}\right)$ for $j=1,\ldots,n$. So that $\mathrm{T}_n\left(\zeta_j^{(n)}\right)=0$. Let $\theta_j^{(n)}=\frac{(2j-1)\pi}{2n}$ so that $\zeta_j^{(n)}=\cos\left(\theta_j^{(n)}\right)$. If $\zeta_j^{(n)}$ is also a zero of $\mathrm{T}_{n+1}$ then $\mathrm{T}_{n+1}\left(\zeta_j^{(n)}\right)=\cos\left((n+1)\theta_j^{(n)}\right)=0$. This implies that $\theta_j^{(n)}$ is a multiple of $\pi$. However, by definition, this is not the case. So $\zeta_j^{(n)}$ cannot also be a zero of $\mathrm{T}_{n+1}$. Therefore, for any $x\in[-1,1]$, if $\mathrm{T}_n(x)=0$, then $\mathrm{T}_{n+1}(x)\ne0$ and the result follows.
\end{enumerate}
\begin{flushright}$\blacksquare$\end{flushright}

Since we are solving the problem in the Cartesian $m$-space identified by the $m$ sample
points, we must prove that the subspace identified by the set of $\mathbf{\tau}_n$, where $n>p$, contains term
vectors a linear combination of which will yield a vector arbitrarily close to the error vector.

\noindent\textbf{Proof}\\
Since the term vectors are derived from the Chebyshev polynomials and the subspace $P$ is dense in $\mathrm{C}(X)$, it follows that there is a linear combination of Chebyshev polynomials in this subspace that is
arbitrarily close to a function passing through the points of the error vector. This will identify a linear
combination of term vectors that is arbitrarily close to the error vector. There will therefore always be at
least one more term vector that reduces the magnitude of the error vector.
\begin{flushright}$\blacksquare$\end{flushright}

As for the rate of convergence, it is a well-known result in approximation theory that if $m$ is
the number of sample points, good results can be obtained by placing your sample points at the zeroes of
$\mathrm{T}_{m+1}$. This yields $m$ orthogonal term vectors. For these points, the CVB approximation algorithm yields the same solution as the CVB interpolation algorithm. Furthermore, the quality of the approximation is consistent with the shape fitting properties of the  CVB approximation algorithm. 

Since we were using noisy data, we did not exploit this result in our application. Instead, we used heavy sampling of the particular area of interest (the region of the workspace that
the robot can actually reach) with minimal sampling of points at the limits of the x and y intervals (i.e.
the corners of the rectangular workspace).
In general, because the norm of the residual error vector decreases as new terms are fitted, and in
most cases only a component of the error vector is removed with each term visited, convergence slows
down as the algorithm progresses. 

\section{The bivariate CVB approximation algorithm}\label{sect:bivariate}

\noindent We use a bivariate version of the CVB approximation algorithm to map image pixels onto object locations.
The bivariate approximation problem is defined as follows:
\begin{definition}[The bivariate polynomial approximation problem]\label{defn:bpa}
Find a polynomial $\mathrm{P}(x,y)$ that fits a set of $m$ sample points $(x_i,y_i,z_i)$; where $i=1,2,\ldots,m$ and the ordered pairs $(x_i,y_i)$ are distinct; such that:
\begin{displaymath}
\max_{i=1}^m|z_i-\mathrm{P}(x_i,y_i)| \le \epsilon, \mathrm{ for\ } \epsilon\ge 0
\end{displaymath}
\end{definition}
Thus:
\begin{equation}\label{eq:bP}
\mathrm{P}(x,y)=\sum_{i=0}^{n-1}\sum_{j=0}^{n-1}a_{i,j}\mathrm{T}_i(x)\mathrm{T}_j(y) 
\end{equation}
where $\mathrm{T}_k$ is the (k+1)'th Chebyshev polynomial of the first kind; $x,y\in[-1,1]$; $a_{i,j}=0$ if $i+j\ge n$ (i.e. the
coefficients form a triangular array). As defined, $\mathrm{P}(x,y)$ is a bivariate polynomial of degree $n-1$.

Casting this problem in Cartesian vector space, we define the bivariate CVB polynomial approximation problem as follows:
\begin{definition}[The bivariate CVB polynomial approximation problem]\label{defn:cvbbpa}
	Given 
	\begin{itemize}
	\item a set of points, $(x_i,y_i,z_i)\in \mathbb{R}^3$, $i = 1,\ldots,m$;
	\item a set of $m$-dimensional Cartesian vectors, $\mathbf{\tau}_{j,k}$, for $j+k<n$ and $j,k=0,\ldots,n-1$, such that the $i$'th component of $\mathbf{\tau}_{j,k}$ is equal to $\mathrm{T}_j(x_i)\mathrm{T}_k(y_i)$, for $i=1,\ldots,m$;
	\item a Cartesian vector, $\mathbf{\gamma}$ such that the $i$'th component of $\mathbf{\gamma}$ is equal to $z_i$, for $i=1,\ldots,m$;
	\item a real value $\epsilon$ ($\epsilon>0$);
	\end{itemize}
	find values for a set of scalar quantities, $a_{j,k}$, such that
	\begin{itemize}
	\item $\mathbf{\rho}=\sum a_{j,k}\mathbf{\tau}_{j,k}$ for $j+k<n$ and $j,k=0,\ldots,n-1$;
	\item $\mathbf{\delta}=\mathbf{\gamma} - \mathbf{\rho}$; 
	\item $|\delta^i|\le \epsilon$ for each component, $\delta^i$, of $\mathbf{\delta}$ ($i=1,\ldots,m$).
	\end{itemize}
\end{definition}
So that the bivariate CVB polynomial approximation problem involves a search for a linear combination of $m$-dimensional Cartesian vectors; where $m$ is the number of sample points. Once the problem has been cast as a CVB problem, the coefficients of the solution are found through a progressive reduction of the residual error vector as for the univariate CVB polynomial approximation problem.

In the univariate version of the CVB approximation algorithm, terms
were visited in order of increasing degree of the corresponding Chebyshev polynomial. In the bivariate case, it is not immediately obvious how to define the order of preference of terms, since several terms can be of the same degree. The following preferencing
rules produce acceptable results for our application:

Let $\langle i,j\rangle$ denote the term with coefficient $a_{i,j}$:
\begin{itemize}
\item The term $\langle a,b\rangle$ is visited before term $\langle c,d\rangle$ if $a+b < c+d$. That is, $i+j$ is used as a primary measure of the ``preference'' for fitting one term over another.
\item If $a+b = c+d$, then $\langle a,b\rangle$ is visited before term $\langle c,d\rangle$ if $\min(a,b) < \min(c,d)$. That is, $\min(i,j)$ is used as a secondary heuristic preference metric.
\item If $a+b = c+d$ and $\min(a,b) = \min(c,d)$, then
$\langle a,b\rangle$ is visited before term $\langle c,d\rangle$ if $a < c$. This is an arbitrary ordering rule to sequence terms of equal preference.
\item After term $\langle i,j\rangle$ is visited, only terms $\langle a,b\rangle$ where $a \le i$ and $b \le j$ are revisited and revisits take place in reverse order to visits.
\end{itemize}
Informally, when visiting, these rules give preference to a term that is of lower degree in the first instance, or if of equal degree, has a component that is of lower degree than any component of a lesser preferred term. When revisiting, the same preferences are observed, but revisits are restricted to terms with components no greater than the corresponding components of the last visited term. So that in general, preference is given to lower order terms. Proof of convergence is similar to that for the univariate case given in section \ref{sect:convergence}.

\subsection{Using the bivariate CVB approximation algorithm}\label{sect:use}

\noindent Fig. \ref{fig:misaligned} (a) shows a deliberately severe misalignment of the camera. The image of the workspace appears to be rotated in a clockwise direction. The camera also produces a pincushion distortion that is reportedly imperceptible to most observers but is significant with respect to locating objects on the workspace. If uncorrected, distortion results in an error of as much as eight millimetres in the location of an object (corresponding to a misplacement of four pixels in the location of an image point). 

In comparison, Fig. \ref{fig:misaligned} (b) shows the results of using the bivariate CVB approximation algorithm to rectify the image
shown in Fig. \ref{fig:misaligned} (a). The rectified image shows a rectangular border to the workspace with edges that are better aligned horizontally and vertically, as can be seen by comparing their alignment with the superimposed grid. In practice, the misalignment will also be imperceptible, but both distortion and misalignment are sufficient to cause errors in locating objects on the workspace.

No attempt was made to ensure that the key points in the test pattern were evenly spaced, or placed according to the zeros of a Chebyshev polynomial. Instead, all but two of the key points\footnote{The key points are the centres of the solid circles in the test pattern shown in Fig. \ref{fig:misaligned}.} were used to sample the region of the workspace that is within reach of the robot. The two extra points were used primarily to fix the corners of the workspace for illustrative purposes. 

An even spacing of the key points in the test pattern is of questionable utility since this does not guarantee an even spacing of their images due to distortion and misalignment. Fig. \ref{fig:noisyline} illustrates the stability of approximations as compared to interpolations when unevenly spaced points are used. This characteristic of approximations was exploited here.

As for using the zeroes of a Chebyshev polynomial, there is the question of which Chebyshev polynomial to use, since, for the CVB approximation algorithm, it is not predetermined how many Chebyshev terms will be included in the approximation. For the rectification depicted in Fig. \ref{fig:misaligned}, the CVB approximation algorithm yielded a solution with $27$ bivariate Chebyshev terms. This yields a polynomial of degree $7$. A bivariate polynomial of degree $n$ can have up to $\frac{1}{2}(n+1)(n+2)$ terms. Given that twenty sample points were used and a bivariate polynomial of degree $20^2=400$ can have up to $80601$ terms, this represents a significant saving in computational overhead.

\section{Summary and conclusions}\label{sect:summary}

\noindent We have developed what we believe to be a novel algorithm for global polynomial approximation. We call this algorithm the Cartesian Vector Based approximation algorithm, or CVB approximation algorithm. This algorithm has several desirable features:
\begin{itemize}
	\item It does not require one to fix the degree of the approximating polynomial ahead of time. The algorithm is capable of progressively adding polynomial terms until the required precision of fit is achieved (or some specified limit on resources is reached). 
	\item Since the algorithm progressively converges on a ``closest fitting function'', an approximate solution is available after each iteration. The longer the algorithm runs, the better the fit.
	\item The algorithm yields better results on the type of data that usually presents difficulties for global polynomial interpolation.
\end{itemize}
It has been proven that the algorithm will yield an approximation that is within $\epsilon$ of a perfect fit, where $\epsilon>0$ and the uniform norm is used as the distance metric. The rate of convergence depends on the choice of data points.

Our focus has been on presenting the CVB approximation algorithm and we use image rectification to illustrate its use. We show how global polynomial approximation can be used to calibrate the vision component of a visually guided pick-and-place robot. Calibration problems are sufficiently prevalent within the field of robotics to render this example relevant.

There are, however, other types of mappings within the field of robotics to which the algorithm may not be directly applicable. For instance, in time series analysis a study is made of a time-varying information-carrying signal for the purpose of predicting its future behaviour. Since time series analysis attempts to predict future events, there is an emphasis on extrapolation. Our algorithm is based on interpolation rather than extrapolation. It is an open question whether it can be adapted for time series analysis. 

Systems identification is another area that uses function approximation \cite{ljung1999}. For instance, ARMAX/NARMAX models are parametrised models of systems consisting of time-varying input values and output values, where the assumption is that the output values depend in some way on current and past input values. System identification involves finding an instantiation of the parameters  that yields a predictor of system behaviour. If a parametrised system model can be contrived that is polynomial in form with the parameters appearing as coefficients, and the data can be transformed to represent points on this polynomial, then any polynomial approximation method (including ours) may be used to estimate the parameters. 

In a case where several approximations exist, there is the question of how the characteristics of the approximation relate to the adequacy of the resulting system model. More specifically, it may be necessary to minimise some domain specific cost function that includes more than the error in the approximation and the contribution of higher order polynomial terms. By limiting the scope of this paper to the details of the CVB approximation algorithm per se, we leave such domain specific questions open for future discussion.

\nocite{ward2005}

\begin{center}
\S\S\S
\end{center}
\bibliographystyle{ieeetr}
\bibliography{gpaareferences}
\end{document}